\relax

\documentclass{article} 
\usepackage{ijcai19}  

\usepackage{times}  
\usepackage{helvet}  
\usepackage{url}  
\usepackage{graphicx}  
\usepackage{amsmath}
\usepackage{amsfonts}
\usepackage{pstricks}
\usepackage{etex}
\usepackage{eso-pic,graphicx}
\usepackage{fancybox}
\usepackage{amsmath,amssymb}
\usepackage{setspace}
\usepackage{xcolor}
\usepackage{array,multirow}
\usepackage{CJK}
\usepackage{tikz}
\usepackage{tikz-qtree}
\usepackage{changepage}
\usepackage{pgfplots}
\usepackage{subfigure}
\usepackage{array} 
\usepackage{bm}

\usepackage{times}
\usepackage{soul}
\usepackage{url}
\usepackage[hidelinks]{hyperref}
\usepackage[utf8]{inputenc}
\usepackage[small]{caption}
\usepackage{graphicx}
\usepackage{amsmath}
\usepackage{booktabs}
\usepackage{algorithm}
\usepackage{algorithmic}
\usetikzlibrary{backgrounds,fit}
\usetikzlibrary{shapes,arrows,shadows}
\usetikzlibrary{patterns}
\usetikzlibrary{arrows,decorations.pathreplacing}
\usetikzlibrary{shadows} 

\usepgflibrary{arrows} 
\usetikzlibrary{arrows} 
\usetikzlibrary{decorations}
\usetikzlibrary{arrows,shapes}

\usetikzlibrary{positioning,fit,calc}

\usetikzlibrary{mindmap,backgrounds} 
\tikzset{
  invisible/.style={opacity=0},
  visible on/.style={alt=#1{}{invisible}},
  alt/.code args={<#1>#2#3}{%
    \alt<#1>{\pgfkeysalso{#2}}{\pgfkeysalso{#3}} 
  },
}

\definecolor{ugreen}{rgb}{0,0.5,0}
\definecolor{lgreen}{rgb}{0.9,1,0.8}
\definecolor{xtgreen1}{rgb}{0.824,0.898,0.8}
\definecolor{xtgreen}{rgb}{0.914,0.945,0.902}
\definecolor{lightgray}{gray}{0.85}
\definecolor{dblue}{cmyk}{0.99998,1,0,0 }
\definecolor{ublue}{rgb}{0.152,0.250,0.545}
\definecolor{myblack}{rgb}{0.15,0.15,0.15}

\newcommand{\PreserveBackslash}[1]{\let\temp=\\#1\let\\=\temp}
\newcolumntype{C}[1]{>{\PreserveBackslash\centering}p{#1}}
\newcolumntype{R}[1]{>{\PreserveBackslash\raggedleft}p{#1}}
\newcolumntype{L}[1]{>{\PreserveBackslash\raggedright}p{#1}}

\usepackage{collcell}
\newcommand*{\MinNumber}{0}%
\newcommand*{\MaxNumber}{1}%
\newcommand{\ApplyGradient}[1]{%
	\pgfmathsetmacro{\PercentColor}{100.0*(#1-\MinNumber)/(\MaxNumber-\MinNumber)}
	\hspace{-0.33em}\colorbox{white!\PercentColor!myblack}{}
}
\newcolumntype{Q}{>{\collectcell\ApplyGradient}c<{\endcollectcell}}

\makeatletter
\newcommand{\distance}[3]{
\tikz@scan@one@point\pgfutil@firstofone($#1-#2$)\relax
\pgfmathsetmacro{#3}{round(0.99626*veclen(\the\pgf@x,\the\pgf@y)/0.0283465)/1000}
}

\newdimen\X
\newdimen\Y
\newdimen\Z
\newdimen\K
\makeatother

\begin{document}
%

\newlength{\wseg}
\newlength{\hseg}
\newlength{\wnode}
\newlength{\hnode}

\setlength{\wseg}{1.5cm}
\setlength{\hseg}{0.6cm}
\setlength{\wnode}{2.5cm}
\setlength{\hnode}{1.1cm}

\definecolor{B1}{RGB}{0,0,0}
\definecolor{B2}{RGB}{20,20,20}
\definecolor{B3}{RGB}{40,40,40}
\definecolor{B4}{RGB}{60,60,60}
\definecolor{B5}{RGB}{80,80,80}
\definecolor{B6}{RGB}{100,100,100}
\definecolor{B7}{RGB}{120,120,120}
\definecolor{B8}{RGB}{140,140,140}
\definecolor{B9}{RGB}{160,160,160}
\definecolor{B10}{RGB}{180,180,180}
\definecolor{white}{RGB}{255,255,255}
\definecolor{black}{RGB}{0,0,0}
\definecolor{blue}{RGB}{0,0,255}
\definecolor{dred}{RGB}{176,7,14}
\definecolor{dblue}{RGB}{0,51,113}
\definecolor{dgreen}{RGB}{0,100,0}

\title{Sharing Attention Weights for Fast Transformer}
\author{
Tong Xiao$^{1,2}$\and
Yinqiao Li$^1$\and
Jingbo Zhu$^{1,2}$\and
Zhengtao Yu$^3$ \textnormal{and}
Tongran Liu$^4$
\affiliations
$^1$Northeastern University, Shenyang, China\\
$^2$NiuTrans Co., Ltd., Shenyang, China\\
$^3$Kunming University of Science and Technology, Kunming, China\\
$^4$CAS Key Laboratory of Behavioral Science, Institute of Psychology, CAS, Beijing, China\\
\emails
\{xiaotong, zhujingbo\}@mail.neu.edu.cn,
\{li.yin.qiao.2012, ztyu\}@hotmail.com,
liutr@psych.ac.cn
}
\maketitle
\begin{abstract}
Recently, the Transformer machine translation system has shown strong results by stacking attention layers on both the source and target-language sides. But the inference of this model is slow due to the heavy use of dot-product attention in auto-regressive decoding. In this paper we speed up Transformer via a fast and lightweight attention model. More specifically, we share attention weights in adjacent layers and enable the efficient re-use of hidden states in a vertical manner. Moreover, the sharing policy can be jointly learned with the MT model. We test our approach on ten WMT and NIST OpenMT tasks. Experimental results show that it yields an average of 1.3X speed-up (with almost no decrease in BLEU) on top of a state-of-the-art implementation that has already adopted a cache for fast inference. Also, our approach obtains a 1.8X speed-up when it works with the \textsc{Aan} model. This is even 16 times faster than the baseline with no use of the attention cache.
\end{abstract}

\section{Introduction}

In recent years, neural models have led to great improvements in machine translation (MT). Approaches of this kind make it possible to learn good mappings between sequences via deep networks and attention mechanisms \cite{sutskever2014sequence,bahdanau2014neural,luong2015effective}. Recent work has explored an architecture that just consists of stacked attentive and feed-forward networks (call it Transformer) \cite{vaswani2017attention}. It makes use of multi-layer dot-product attention to capture the dependency among language units. Beyond this, training this kind of model is fast because we can parallelize computation over all positions of the sequence on modern graphics processing units (GPUs). These properties make Transformer popular in recent MT evaluations and industrial deployments.

However, standard implementations of Transformer are prone to slow inference though fast in training. At test time, the system produces one target word each time until an end symbol is reached. This process is auto-regressive and slow because we have to run dot-product attention for each position rather than batching the computation of the sequence. The situation is even worse if 6 or more attention layers are stacked and the attention model occupies the inference time. To address this issue, efficient networks have been investigated. For example, one can replace dot-product attention with additive attention and use average attention models instead  \cite{Zhang2018AAN}, or explore non-autoregressive decoders that benefit from the trick of batched matrix operations over the entire sequence \cite{Gu2017NonAutoregressive}. But these methods either lose the explicit model of word dependencies, or require complicated networks that are hard to train.

In this work, we observe that the attention model shares a similar distribution among layers in weighting different positions of the sequence. This experience lead us to study the issue in another line of research, in which we reduce redundant computation and re-use some of the hidden states in the attention network. We propose a method to share attention weights in adjacent layers (call it shared attention network, or \textsc{San} for short). It leads to a model that shares attention computation in the stacked layers vertically. In addition to the new architecture, we develop a joint method to learn sharing policies and MT models simultaneously. As another ``bonus", \textsc{San} reduces the memory footprint because some hidden states are kept in the same piece of memory.

\textsc{San} is simple and can be implemented in a few hours by anyone with an existing kit of Transformer. Also, it is orthogonal to previous methods and is straightforwardly applicable to the variants of Transformer. We test our approach in a state-of-the-art system where an attention cache is already in use for speed-up. Experimental results on ten WMT and NIST OpenMT tasks show an average of 1.3X speed-up with almost no decrease in BLEU. More interestingly, it obtains a bigger speed-up (1.8X) when working with the \textsc{Aan} model. The best result is 16 times faster than the baseline where no cache is adopted.

\section{The Transformer System}

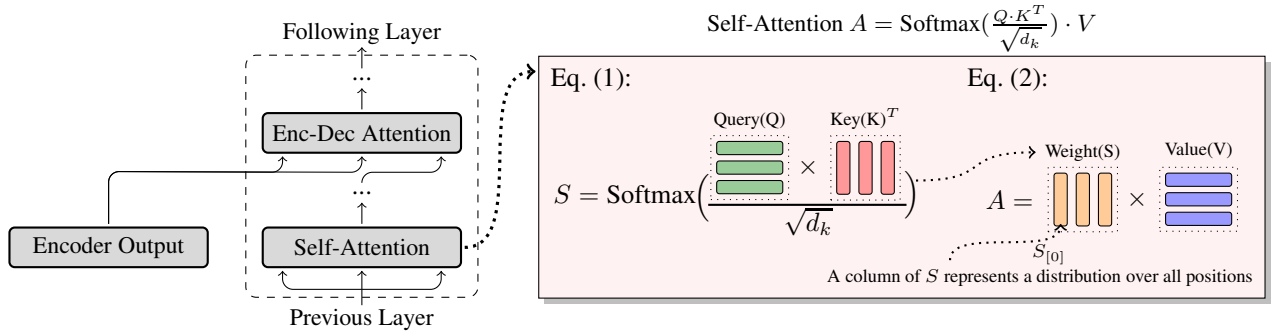
\begin{figure*}[!htb]
\begin{center}
\setlength{\tabcolsep}{1pt}
\begin{tikzpicture}[decoration=brace]
\begin{scope}
\tikzstyle{sublayernode} = [rectangle,draw,thick,inner sep=3pt,rounded corners=2pt,align=center,minimum width=7.5em,minimum height=1.5em,font=\footnotesize]
\tikzstyle{inputnode} = [rectangle,inner sep=3pt,align=center,minimum width=\wnode,font=\footnotesize]
\tikzstyle{labelnode} = [rotate=90]

\node[anchor=south,sublayernode,fill=gray!30] (satt) at (0, 0) {Self-Attention};
\node[anchor=south,inner sep=2pt] (connect01) at ([yshift=1.0em]satt.north) {...};
\node[anchor=south,sublayernode,fill=gray!30] (edatt) at ([yshift=1.2em]connect01.north) {Enc-Dec Attention};
\node[anchor=south,inner sep=2pt] (connect02) at ([yshift=1.0em]edatt.north) {...};

\node[anchor=north,inner sep=2pt] (deinput) at ([yshift=-1.3em]satt.south) {\footnotesize{Previous Layer}};
\node[anchor=south,inner sep=2pt] (deoutput) at ([yshift=0.8em]connect02.north) {\footnotesize{Following Layer}};

\node[anchor=east,sublayernode,fill=gray!30] (encoder) at ([xshift=-2em]satt.west) {Encoder Output};

\draw[->] ([yshift=0.1em]satt.north) --  ([yshift=-0.1em]connect01.south);
\draw[->] ([yshift=0.1em]edatt.north) --  ([yshift=-0.1em]connect02.south);
\draw[->,rounded corners] ([yshift=0.1em]connect01.north) -- ([yshift=0.5em]connect01.north) -- ([yshift=0.5em,xshift=3em]connect01.north) -- ([yshift=-0.1em,xshift=3em]edatt.south);
\draw[->,rounded corners] ([yshift=0.1em]encoder.north) -- ([yshift=2.08em]encoder.north) -- ([yshift=0.5em]connect01.north) -- ([yshift=-0.1em]edatt.south);
\draw[->,rounded corners] ([yshift=0.1em]encoder.north) -- ([yshift=2.08em]encoder.north) -- ([yshift=0.5em,xshift=-3em]connect01.north) -- ([yshift=-0.1em,xshift=-3em]edatt.south);
\draw[->,rounded corners] ([yshift=0.5em]deinput.north) -- ([yshift=0.5em,xshift=3em]deinput.north) -- ([yshift=-0.1em,xshift=3em]satt.south);
\draw[->,rounded corners] ([yshift=0.5em]deinput.north) -- ([yshift=0.5em,xshift=-3em]deinput.north) -- ([yshift=-0.1em,xshift=-3em]satt.south);
\draw[->,rounded corners] ([yshift=-0.1em]deinput.north) -- ([yshift=-0.1em]satt.south);
\draw[->,rounded corners] ([yshift=0.1em]connect02.north) -- ([yshift=0.1em]deoutput.south);

\begin{pgfonlayer}{background}
    \coordinate (debottom) at ([yshift=0.8em]deinput.north);
    \node[rectangle,draw,dashed,inner sep=6pt,rounded corners] [fit = (satt) (connect02) (debottom)] (decoder) {};
\end{pgfonlayer}

\end{scope}

\begin{scope}
\tikzstyle{rownode} = [rectangle,draw,thin,inner sep=0pt,rounded corners=1pt,align=center,minimum width=2.5em,minimum height=0.5em,font=\footnotesize]
\tikzstyle{colnode} = [rectangle,draw,thin,inner sep=0pt,rounded corners=1pt,align=center,minimum width=0.5em,minimum height=2.0em,font=\footnotesize]

\node[anchor=west] (eq01) at ([xshift=3em,yshift=2em]edatt.east) {Eq. (1):};
\node[anchor=west] (eq02) at ([xshift=12em]eq01.east){Eq. (2):};

\node[anchor=north,rownode,fill=ugreen!40] (query01) at ([yshift=-1.5em,xshift=6em]eq01.south) {};
\node[anchor=north,rownode,fill=ugreen!40] (query02) at ([yshift=-0.2em]query01.south) {};
\node[anchor=north,rownode,fill=ugreen!40] (query03) at ([yshift=-0.2em]query02.south) {};

\node[anchor=north west,colnode,fill=red!40] (key01) at ([xshift=2em]query01.north east) {};
\node[anchor=west,colnode,fill=red!40] (key02) at ([xshift=0.3em]key01.east) {};
\node[anchor=west,colnode,fill=red!40] (key03) at ([xshift=0.3em]key02.east) {};

\node[anchor=south] (querylabel) at ([yshift=0.1em]query01.north) {\scriptsize{Query(Q)}};
\node[anchor=south] (keylabel) at ([yshift=0.1em]key02.north) {\scriptsize{Key(K)$^T$}};
\node[anchor=west] (timeslabel) at ([xshift=0.2em]query02.east) {$\times$};
\node[anchor=east] (statelabel) at ([xshift=-0.4em]query03.south west) {$S=\textrm{Softmax}$};
\node[anchor=north] (dlabel) at ([yshift=-0.5em]timeslabel.south) {$\sqrt{d_k}$};

\node[anchor=east] (braceleft) at ([yshift=-0.4em,xshift=0.1em]query03.west) {$\Big($};
\node[anchor=west] (braceright) at ([yshift=-0.0em,xshift=0.0em]key03.south east) {$\Big)$};
\draw[-,thick] ([yshift=-0.4em,xshift=-0.3em]query03.south west) -- ([yshift=-0.4em,xshift=7.2em]query03.south west);

\node[anchor=north west,colnode,fill=orange!40] (state01) at ([xshift=6em,yshift=-1.2em]key03.north east) {};
\node[anchor=west,colnode,fill=orange!40] (state02) at ([xshift=0.3em]state01.east) {};
\node[anchor=west,colnode,fill=orange!40] (state03) at ([xshift=0.3em]state02.east) {};

\node[anchor=north west,rownode,fill=blue!40] (value01) at ([yshift=0em,xshift=2em]state03.north east) {};
\node[anchor=north,rownode,fill=blue!40] (value02) at ([yshift=-0.2em]value01.south) {};
\node[anchor=north,rownode,fill=blue!40] (value03) at ([yshift=-0.2em]value02.south) {};

\begin{pgfonlayer}{background}

\end{pgfonlayer}

\node[anchor=south] (valuelabel) at ([yshift=0.2em]value01.north) {\scriptsize{Value(V)}};
\node[anchor=south] (slabel) at ([yshift=0.1em]state02.north) {\scriptsize{Weight(S)}};
\node[anchor=west] (timeslabel02) at ([xshift=0.2em]state03.east) {$\times$};
\node[anchor=east] (attlabel) at ([xshift=-0.4em]state01.west) {$A=$};
\node[anchor=north west] (weightlabel) at ([yshift=-1.6em,xshift=-6em]attlabel.south west) {\scriptsize{A column of $S$ represents a distribution over all positions}};

\draw[->,dotted,thick] ([xshift=5em]weightlabel.north west).. controls +(north:1.3em) and +(south:1.3em) .. ([yshift=0.1em]state01.south) node[anchor=north,pos=0.8,yshift=0.1em]  {\scriptsize{$S_{[0]}$}};
\draw[->,dotted,thick] ([xshift=0.8em,yshift=-1.5em]key03.north east).. controls +(east:3em) and +(west:3em) .. ([xshift=-0.0em,yshift=-0.0em]slabel.west);

\begin{pgfonlayer}{background}
    \coordinate (valueright) at ([xshift=0.3em]value02.east);
    \node[rectangle,draw,inner sep=1pt,drop shadow,fill=red!5] [fit = (eq01) (value03) (weightlabel) (valueright)] (attdescription) {};
    \node[rectangle,draw,dotted,inner sep=2pt] [fit = (query01) (query02) (query03)] (query) {};
    \node[rectangle,draw,dotted,inner sep=2pt] [fit = (key01) (key02) (key03)] (key) {};
    \node[rectangle,draw,dotted,inner sep=2pt] [fit = (state01) (state02) (state03)] (state) {};
    \node[rectangle,draw,dotted,inner sep=2pt] [fit = (value01) (value02) (value03)] (value) {};
\end{pgfonlayer}

\node[anchor=south] (selfattlabel) at ([yshift=-0.2em]attdescription.north) {\footnotesize{Self-Attention $A=\textrm{Softmax}(\frac{Q \cdot {K}^T}{\sqrt{d_k}}) \cdot V$}};

\draw[->,very thick,dotted] ([xshift=0.1em]satt.east) .. controls +(east:2.5em) and +(west:2.5em) .. ([yshift=4em,xshift=-0.1em]attdescription.west);

\end{scope}
\end{tikzpicture}
\end{center}
\caption{\fontsize{9pt}{10.8pt}\selectfont Decoder-side attention sub-layers in Transformer}
\label{fig:transformer}
\end{figure*}

The Transformer system follows the popular encoder-decoder paradigm. On the encoder side, there are a number of identical stacked layers. Each of them is composed of a self-attention sub-layer and a feed-forward sub-layer. In Transformer, the attention model is scaled dot-product attention. Let $l$ be the length of the source sequence. The input of the attention sub-layer is a tuple of $(Q,K,V)$, where $Q \in \mathbb{R}^{l \times d_k}$, $K \in \mathbb{R}^{l \times d_k}$ and $V \in \mathbb{R}^{l \times d_v}$ are the matrices of queries, keys, and values packed over the sequence. In self-attention, we first compute the dot-product of queries and keys, followed by the rescaling and softmax operations.

\begin{equation}
S=\textrm{Softmax}(\frac{Q \cdot {K}^T}{\sqrt{d_k}}) \label{math-dot}
\end{equation}

\noindent $S$ is an $l \times l$ matrix, where entry $(i,j)$ represents the strength of connecting position $i$ with position $j$. Note that $S$ is essentially a weight (or scalar) matrix where every column represents a distribution. The output of self-attention is simply defined as the weighted sum of values:

\begin{equation}
A=S \cdot V \label{math-att}
\end{equation}

\noindent Here $Q$, $K$ and $V$ are generated from the same source with a linear transformation. The self-attention result is then fed into a fully connected feed-forward network (FFN).

The decoder shares a similar structure with the encoder. Apart from the self-attention sub-layer, an encoder-decoder attention sub-layer is introduced to model the correspondence between source positions and target positions. Basically, the encoder-decoder attention has the same form as Eqs. (\ref{math-dot}) and (\ref{math-att}), where the queries come from the output of the previous layer, and the keys and values come from the output of the encoder. See Figure \ref{fig:transformer} for an illustration of the attention model used in Transformer.

Note that the matrix multiplications in Eqs. (\ref{math-dot}) and (\ref{math-att}) are time consuming. It is a bigger problem for inference because Eqs. (\ref{math-dot}) and (\ref{math-att}) repeat for each position until we finish the generation.

\section{Shared Attention Networks}

In this work we speed up the decoder-side attention because the decoder is the heaviest component in Transformer.

\subsection{Attention Weights}

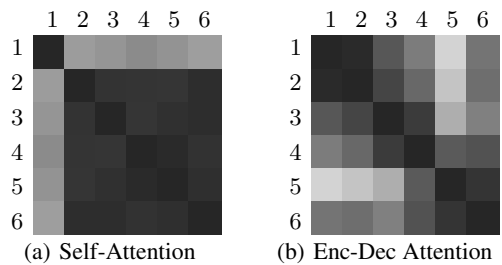
\begin{figure}[t]
	\begin{center}
		\renewcommand{\arraystretch}{0}
		\setlength{\tabcolsep}{0.5mm}
		\setlength{\fboxsep}{2.2mm} 
		\begin{tabular}{C{.20\textwidth}C{.20\textwidth}C{.20\textwidth}C{.20\textwidth}}
			\setlength{\tabcolsep}{0pt}
			\subfigure [\footnotesize{Self-Attention}] {
				\begin{tabular}{cc}
					\setlength{\tabcolsep}{0pt}
					~
					&
					\begin{tikzpicture}
					\begin{scope}						
					\node [inner sep=1.5pt] (w1) at (0,0) {\small{$1$} };
					
					\foreach \x/\y/\z in {2/1/$2$, 3/2/$3$, 4/3/$4$, 5/4/$5$, 6/5/$6$}
					{
						\node [inner sep=1.5pt,anchor=south west] (w\x) at ([xshift=1.15em]w\y.south west) {\small{\z} };
					}
					\end{scope}
					\end{tikzpicture}
					\\
					\renewcommand\arraystretch{1.2}
					\begin{tabular}{c}
						\setlength{\tabcolsep}{0pt}
						\small{$1\ \ $} \\
						\small{$2\ $} \\
						\small{$3\ $} \\
						\small{$4\ $} \\
						\small{$5\ $} \\
						\small{$6\ $} \\
					\end{tabular}
					&
					\begin{tabular}{*{6}{Q}}
						0.0000 & 0.5429 & 0.5138 & 0.4650 & 0.5005 & 0.5531 \\

                        0.5429 & 0.0000 & 0.0606 & 0.0630 & 0.0703 & 0.0332 \\

                        0.5138 & 0.0606 & 0.0000 & 0.0671 & 0.0472 & 0.0296 \\

                        0.4650 & 0.0630 & 0.0671 & 0.0000 & 0.0176 & 0.0552 \\

                        0.5005 & 0.0703 & 0.0472 & 0.0176 & 0.0000 & 0.0389 \\

                        0.5531 & 0.0332 & 0.0296 & 0.0552 & 0.0389 & 0.0000 \\
					\end{tabular}
				\end{tabular}
			}
			&
			
			\subfigure [\footnotesize{Enc-Dec Attention}] {
				\setlength{\tabcolsep}{0pt}
				\begin{tabular}{cc}
					\setlength{\tabcolsep}{0pt}
					~
					&
					\begin{tikzpicture}
					\begin{scope}						
					\node [inner sep=1.5pt] (w1) at (0,0) {\small{$1$} };
					\foreach \x/\y/\z in {2/1/$2$, 3/2/$3$, 4/3/$4$, 5/4/$5$, 6/5/$6$}
					{
						\node [inner sep=1.5pt,anchor=south west] (w\x) at ([xshift=1.15em]w\y.south west) {\small{\z} };
					}
					\end{scope}
					\end{tikzpicture}
					\\
					\renewcommand\arraystretch{1.2}
					\begin{tabular}{c}
						\setlength{\tabcolsep}{0pt}
						\small{$1\ \ $} \\
						\small{$2\ $} \\
						\small{$3\ $} \\
						\small{$4\ $} \\
						\small{$5\ $} \\
						\small{$6\ $} \\
					\end{tabular}
					&
					\begin{tabular}{*{6}{Q}}
                        0.0000 & 0.0175 & 0.2239 & 0.3933 & 0.7986 & 0.3603 \\

                        0.0175 & 0.0000 & 0.1442 & 0.3029 & 0.7295 & 0.3324 \\

                        0.2239 & 0.1442 & 0.0000 & 0.0971 & 0.6270 & 0.4163 \\

                        0.3933 & 0.3029 & 0.0971 & 0.0000 & 0.2385 & 0.2022 \\

                        0.7986 & 0.7295 & 0.6270 & 0.2385 & 0.0000 & 0.0658 \\

                        0.3603 & 0.3324 & 0.4163 & 0.2022 & 0.0658 & 0.0000 \\
					\end{tabular}
				\end{tabular}
			}
		\end{tabular}
	\end{center}
	
	\begin{center}
		\caption{\fontsize{9pt}{10.8pt}\selectfont The Jensen-Shannon divergence of the attention weights for every pair of layers on the WMT14 English-German task (a dark cell means the distributions are similar)}
		\label{figure:js}
	\end{center}
\end{figure}

Self-attention is essentially a procedure that fuses the input values to form a new value at each position. Let $S_{[i]}$ be column $i$ of weight matrix $S$. For position $i$ , we first compute $S_{[i]}$ to weight all positions (as in Eq. (\ref{math-dot})), and then compute the weighted sum of values by $S_{[i]}$ (as in Eq. (\ref{math-att})).  In column vector $S_{[i]}$, element $S_{i,j}$ indicates the contribution that we fuse the value at position $j$ to position $i$. Intuitively, the attention weight $S_{[i]}$ should not be volatile in different levels of language representation because the correlations between positions somehow reflect the dependency of language units. For example, for an English sentence, the subject and the verb correlate well no matter how many layers we make on top of the input sequence. On the other hand, the subject and the adverbial may not have a big impact to each other in all stacked layers.

\begin{figure*}[]
\centering
\begin{tikzpicture}[decoration=brace]
\begin{scope}
\tikzstyle{layernode} = [rectangle,draw,thick,densely dotted,inner sep=3pt,rounded corners,minimum width=2.1\wnode,minimum height=2.7\hnode]
\tikzstyle{attnnode} = [rectangle,draw,inner sep=3pt,rounded corners, minimum width=2\wnode,minimum height=2.2\hnode]
\tikzstyle{thinnode} = [rectangle,inner sep=1pt,rounded corners=1pt,minimum size=0.3\hnode,font=\scriptsize]
\tikzstyle{fatnode} = [rectangle,inner sep=1pt,rounded corners=1pt,minimum height=0.3\hnode,minimum width=\wnode,font=\small]

\coordinate (layer00) at (0,0);
\foreach \i / \j in {1/0,2/1,3/2,4/3,5/4}
    \coordinate (layer0\i) at ([xshift=2.2\wnode+0.3\wseg]layer0\j);

\node[layernode,anchor=north] (layer11) at ([yshift=-\hseg]layer01.south) {};
\node[attnnode,anchor=south] (attn11) at ([yshift=0.1\hnode]layer11.south) {};
\node[anchor=north west,inner sep=4pt,font=\small] () at (attn11.north west) {Attention};
\node[anchor=south,inner sep=0pt] (out11) at ([yshift=0.3\hseg]attn11.north) {$\cdots$};
\node[thinnode,anchor=south west,thick,draw=dblue,text=black] (q11) at ([xshift=0.1\wseg,yshift=0.2\hseg]attn11.south west) {$Q^n$};
\node[thinnode,anchor=south,thick,draw=orange,text=black] (k11) at ([yshift=0.2\hseg]attn11.south) {$K^n$};
\node[thinnode,anchor=south east,thick,draw=purple,text=black] (v11) at ([xshift=-0.1\wseg,yshift=0.2\hseg]attn11.south east) {$V^n$};
\node[fatnode,anchor=south,thick,draw] (s11) at ([xshift=0.5\wseg,yshift=0.8\hseg]q11.north east) {$S^n\!=\!S(Q^n\!\cdot\!K^n)$};
\node[fatnode,anchor=south,thick,draw] (a11) at ([xshift=0.45\wseg,yshift=1.3\hseg+0.6\hnode]k11.north east) {$A^n\!=\!S^n\!\cdot\!V$};
\begin{scope}[fill=black!100]
    \draw[-latex',thick,draw=black!100] (q11.north) .. controls +(north:0.5\hseg) and +(south:0.8\hseg) .. (s11.south);
    \draw[-latex',thick,draw=black!100] (k11.north) .. controls +(north:0.5\hseg) and +(south:0.8\hseg) .. (s11.south);
\end{scope}
\begin{scope}[fill=black!100]
    \draw[-latex',thick,draw=black!100] (s11.north) .. controls +(north:0.7\hseg) and +(south:0.8\hseg) ..(a11.south);
    \draw[-latex',thick,draw=black!100] (v11.north) .. controls +(north:2.7\hseg) and +(south:0.9\hseg) .. (a11.south);
\end{scope}
\draw[-latex',thick] (a11.north).. controls +(north:0.3\hseg) and +(south:0.7\hseg) ..(out11.south);

\node[layernode,anchor=north] (layer12) at ([yshift=-\hseg]layer02.south) {};
\node[attnnode,anchor=south] (attn12) at ([yshift=0.1\hnode]layer12.south) {};
\node[anchor=north west,inner sep=4pt,font=\small] () at (attn12.north west) {Attention};
\node[anchor=south,inner sep=0pt] (out12) at ([yshift=0.3\hseg]attn12.north) {$\cdots$};
\node[thinnode,anchor=south west,thick,draw=dblue!40,text=black!40] (q12) at ([xshift=0.1\wseg,yshift=0.2\hseg]attn12.south west) {$Q^n$};
\node[thinnode,anchor=south,thick,draw=orange!40,text=black!40] (k12) at ([yshift=0.2\hseg]attn12.south) {$K^n$};
\node[thinnode,anchor=south east,thick,draw=purple,text=black] (v12) at ([xshift=-0.1\wseg,yshift=0.2\hseg]attn12.south east) {$V^n$};
\node[fatnode,anchor=south,thick,densely dashed,draw] (s12) at ([xshift=0.5\wseg,yshift=0.8\hseg]q12.north east) {$S^n\!=\!S^m$};
\node[fatnode,anchor=south,thick,draw] (a12) at ([xshift=0.45\wseg,yshift=1.3\hseg+0.6\hnode]k12.north east) {$A^n\!=\!S^n\!\cdot\!V$};
\begin{scope}[fill=black!40]
    \draw[-latex',thick,draw=black!40] (q12.north) .. controls +(north:0.5\hseg) and +(south:0.8\hseg) .. (s12.south);
    \draw[-latex',thick,draw=black!40] (k12.north) .. controls +(north:0.5\hseg) and +(south:0.8\hseg) .. (s12.south);
\end{scope}
\begin{scope}[fill=black!100]
    \draw[-latex',thick,draw=black!100] (s12.north).. controls +(north:0.7\hseg) and +(south:0.8\hseg) .. (a12.south);
    \draw[-latex',thick,draw=black!100] (v12.north).. controls +(north:2.7\hseg) and +(south:0.9\hseg) .. (a12.south);
\end{scope}
\draw[-latex',thick] (a12.north).. controls +(north:0.3\hseg) and +(south:0.7\hseg) ..(out12.south);

\node[layernode,anchor=north] (layer13) at ([yshift=-\hseg]layer03.south) {};
\node[attnnode,anchor=south] (attn13) at ([yshift=0.1\hnode]layer13.south) {};
\node[anchor=north west,inner sep=4pt,font=\small] () at (attn13.north west) {Attention};
\node[anchor=south,inner sep=0pt] (out13) at ([yshift=0.3\hseg]attn13.north) {$\cdots$};
\node[thinnode,anchor=south west,thick,draw=dblue!40,text=black!40] (q13) at ([xshift=0.1\wseg,yshift=0.2\hseg]attn13.south west) {$Q^n$};
\node[thinnode,anchor=south,thick,draw=orange!40,text=black!40] (k13) at ([yshift=0.2\hseg]attn13.south) {$K^n$};
\node[thinnode,anchor=south east,thick,draw=purple!40,text=black!40] (v13) at ([xshift=-0.1\wseg,yshift=0.2\hseg]attn13.south east) {$V^n$};
\node[fatnode,anchor=south,thick,draw=black!40,text=black!40] (s13) at ([xshift=0.5\wseg,yshift=0.8\hseg]q13.north east) {$S^n$};
\node[fatnode,anchor=south,thick,densely dashed,draw] (a13) at ([xshift=0.45\wseg,yshift=1.3\hseg+0.6\hnode]k13.north east) {$A^n\!=\!A^m$};
\begin{scope}[fill=black!40]
    \draw[-latex',thick,draw=black!40] (q13.north) .. controls +(north:0.5\hseg) and +(south:0.8\hseg) .. (s13.south);
    \draw[-latex',thick,draw=black!40] (k13.north) .. controls +(north:0.5\hseg) and +(south:0.8\hseg) .. (s13.south);
\end{scope}
\begin{scope}[fill=black!40]
    \draw[-latex',thick,draw=black!40] (s13.north) .. controls +(north:0.7\hseg) and +(south:0.8\hseg) .. (a13.south);
    \draw[-latex',thick,draw=black!40] (v13.north) .. controls +(north:2.7\hseg) and +(south:0.9\hseg) .. (a13.south);
\end{scope}
\draw[-latex',thick] (a13.north).. controls +(north:0.3\hseg) and +(south:0.7\hseg) ..(out13.south);

\foreach \i / \j / \k / \q / \s / \t / \v in
    {2/1/1/100/100/100/100, 2/2/1/100/100/100/100, 2/3/1/100/100/100/100}
    {
        \node[layernode,anchor=north] (layer\i\j) at ([yshift=-0.8\hseg]layer\k\j.south) {};
        \node[attnnode,anchor=south] (attn\i\j) at ([yshift=0.1\hnode]layer\i\j.south) {};
        \node[anchor=north west,inner sep=4pt,font=\small] () at (attn\i\j.north west) {Attention};
        \node[anchor=south,inner sep=0pt] (out\i\j) at ([yshift=0.3\hseg]attn\i\j.north) {$\cdots$};

        \node[thinnode,anchor=south west,thick,draw=dblue!\q,text=black] (q\i\j) at ([xshift=0.1\wseg,yshift=0.2\hseg]attn\i\j.south west) {$Q^m$};
        \node[thinnode,anchor=south,thick,draw=orange!\q,text=black] (k\i\j) at ([yshift=0.2\hseg]attn\i\j.south) {$K^m$};
        \node[thinnode,anchor=south east,thick,draw=purple!\s,text=black] (v\i\j) at ([xshift=-0.1\wseg,yshift=0.2\hseg]attn\i\j.south east) {$V^m$};
        \node[fatnode,anchor=south,thick,draw=black!\s] (s\i\j) at ([xshift=0.45\wseg,yshift=0.8\hseg]q\i\j.north east) {$S^m\!=\!S(Q^m\!\cdot\!K^m)$};
        \node[fatnode,anchor=south,thick,draw=black!80] (a\i\j) at ([xshift=0.45\wseg,yshift=1.3\hseg+0.6\hnode]k\i\j.north east) {$A^m\!=\!S^m\!\cdot\!V$};
        \begin{scope}[fill=black!\q]
            \draw[-latex',thick,draw=black!\t] (q\i\j.north) .. controls +(north:0.5\hseg) and +(south:0.8\hseg) .. (s\i\j.south);
            \draw[-latex',thick,draw=black!\t] (k\i\j.north) .. controls +(north:0.5\hseg) and +(south:0.8\hseg) .. (s\i\j.south);
        \end{scope}
        \begin{scope}[fill=black!\s]
            \draw[-latex',thick,draw=black!\v] (s\i\j.north).. controls +(north:0.7\hseg) and +(south:0.8\hseg) ..(a\i\j.south);
            \draw[-latex',thick,draw=black!\v] (v\i\j.north).. controls +(north:2.7\hseg) and +(south:0.9\hseg) ..(a\i\j.south);
        \end{scope}
        \draw[-latex',thick] (a\i\j.north).. controls +(north:0.3\hseg) and +(south:0.7\hseg) ..(out\i\j.south);
    }

\draw[-latex',densely dashed,very thick] (s22.west) to [out=120,in=-120] (s12.west);
\draw[-latex',densely dashed,very thick] (a23.east) to [out=60,in=-60] (a13.east);

\foreach \i in {1,2,3}
{
    \node[anchor=north west,inner sep=3pt,font=\small] () at (layer1\i.north west) {Layer $n\!=\!m\!+\!i$};
    \node[anchor=north west,inner sep=3pt,font=\small] () at (layer2\i.north west) {Layer $m$};
    \node[anchor=center,inner sep=1pt] (dot1\i) at ([yshift=0.5\hseg]layer1\i.north) {$\cdots$};
    \draw[->,thick] (out1\i.north) -- ([yshift=0.1em]dot1\i.south);
    \node[anchor=center,inner sep=1pt] (dot2\i) at ([yshift=-0.4\hseg]layer1\i.south) {$\cdots$};
    \draw[->,thick] ([yshift=-0.15em]dot2\i.north) -- ([yshift=-0.3em]attn1\i.south);
    \draw[->,thick] (out2\i.north) -- ([yshift=0.1em]dot2\i.south);
    \node[anchor=center,inner sep=1pt] (dot3\i) at ([yshift=-0.4\hseg]layer2\i.south) {$\cdots$};
    \draw[->,thick] ([yshift=-0.15em]dot3\i.north) -- ([yshift=-0.3em]attn2\i.south);
}

\node[anchor=north,align=left,inner sep=1pt,font=\normalsize] () at (dot31.south) {(a) Standard Transformer Attention};
\node[anchor=north,align=left,inner sep=1pt,font=\normalsize] () at (dot32.south) {(b) \textsc{San} Self-Attention};
\node[anchor=north,align=left,inner sep=1pt,font=\normalsize] () at (dot33.south) {(c) \textsc{San} Encoder-Decoder Attention};

\end{scope}
\end{tikzpicture}
\caption{\fontsize{9pt}{10.8pt}\selectfont Comparison of the standard attention model and the \textsc{San} model}
\label{fig:san}
\end{figure*}
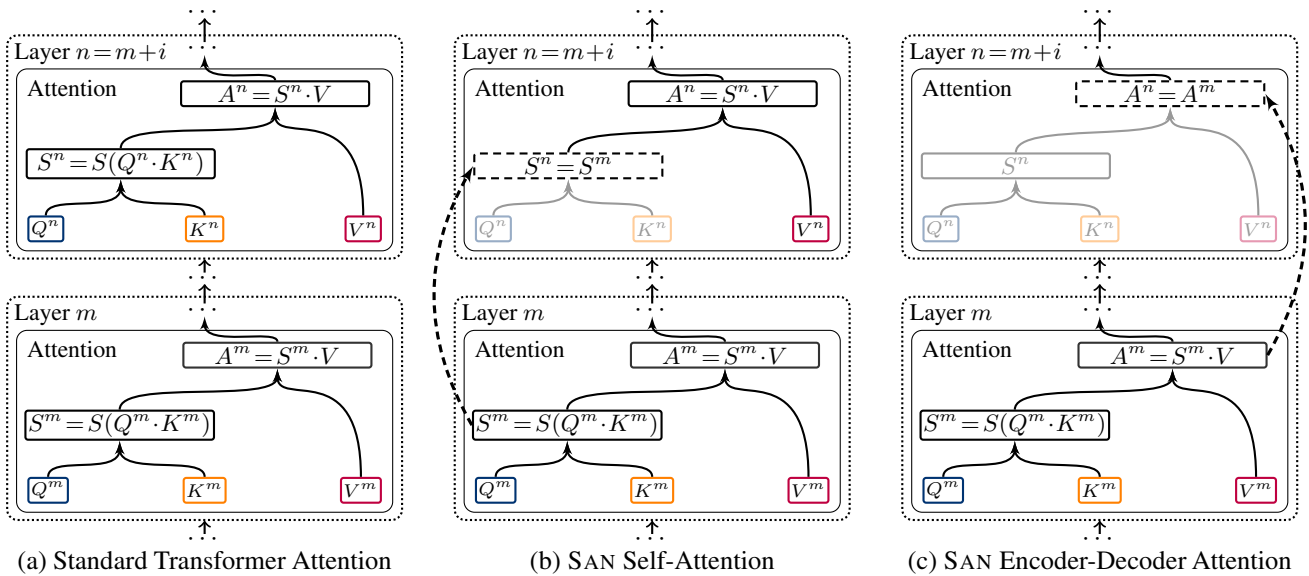

To verify this, we compute the Jensen-Shannon (JS) divergence to measure how the attention weight distribution of a layer is different from another \cite{Jianhua1991JS}. We choose the JS divergence because it is symmetric and bounded. For multi-head attention, we regard different heads as separate channels. We compute the JS score for each individual head and then average them for final output. Figure \ref{figure:js} shows that the system generates similar weights over layers. For self-attention, layers 2-6 almost enjoy the same weight distribution. For encoder-decoder attention, we observe a larger variance but good similarities still exist among two or three adjacent layers (see entries around the diagonal of Figure \ref{figure:js}(b)). All these show the possibility of removing redundant computation in Transformer.

\subsection{The Model}

An obvious next step is to develop a faster attention model that makes efficient re-use of the states in Eqs. (\ref{math-dot}) and (\ref{math-att}), instead of computing everything on the fly. In this work we present a shared attention network (\textsc{San}) to share weight matrix $S$ for adjacent layers. The idea is that we just compute the weight matrix once and reuse it for upper-level layers. Here we describe \textsc{San} for both the self-attention and encoder-decoder attention models.

\begin{itemize}
\item \textbf{\textsc{San} Self-Attention}. We define the self-attention weight matrix in layer $m$ as:

\begin{equation}
S^m=s(Q^m,K^m) \label{math-att-func}
\end{equation}

\noindent where $s(\cdot,\cdot)$ is the function described in Eq. (\ref{math-dot}), $Q^m$ and $K^m$ are the inputs, and $S^m$ is the attention weight for the output. In \textsc{San}, we can share $S^m$ with the layers above $m$, like this

\begin{eqnarray}
S^{m+i} & =  &s(Q^m,K^m) \label{math-att-func} \\
        & \textrm{for} &  i \in [1,\pi-1] \nonumber
\end{eqnarray}

\noindent where $\pi$ indicates how many layers share the same attention weights. For example, in a 6-layer decoder, we can share the self-attention weights for every two layers ($\pi=2$), or  share the weights for the first two layers ($\pi_1=2$) and let the remaining 4 layers use another weights ($\pi_2=4$). We discuss the sharing strategy in the later part of the section.

\item \textbf{\textsc{San} Encoder-Decoder Attention}. For encoder-decoder attention, we do the same thing as in self-attention, but with a trick for further speed-up. In encoder-decoder attention, keys and values are from the output of the encoder, i.e., $K$ and $V$ have already been shared among layers on the decoder side. In response, we can share $A=S \cdot V$ for encoder-decoder attention layers. This can be described as

\begin{eqnarray}
A^{m+i} & =  & A^{m} \nonumber \\
        & =  & S^m \cdot V \label{math-att-ende-func} \\
        & \textrm{for} &  i \in [1,\pi-1] \nonumber
\end{eqnarray}

\noindent where $A^m$ is the attention output of layer $m$, $V$ is the context representation generated by the encoder. See Figure \ref{fig:san} for a comparison of the standard attention model and \textsc{San}.

\end{itemize}

In addition to system speed-up, \textsc{San} also reduces the memory footprint. In \textsc{San}, we just need one data copy of weight matrix for a layer block, rather than allocating memory space for every layer. Moreover, the linear transformation of the input (i.e., $Q$ and $K$) can be discarded when the attention weights come from another layer. It reduces both the number of model parameters and the memory footprint in inference.

Another note on \textsc{San}. \textsc{San} is a process that simplifies the model and re-uses hidden states in the network. It is doing something similar to systems that share model parameters in different levels of the network \cite{wu2016google,Yang2018Unsupervised,Luong2016Multi}. Such methods have been proven to improve the robustness of neural models on many natural language processing tasks. Sharing parameters and/or hidden states can reduce the model complexity. Previous work has pointed out that MT systems cannot benefit a lot from very deep and complex networks \cite{vaswani2017attention,Denny2017Massive}. \textsc{San} might alleviate this problem and makes it easier to train neural MT models. For example, in our experiments, we see that \textsc{San} can improve translation quality in some cases in addition to considerable speed-ups.

\subsection{Learning to Share}
\label{sec:learning-to-share}

The remaining issue is how to decide which layers can be shared. A simple way is to use the same setting of $\pi$ for the entire layer stack, and tune it on a development set. For example, we can try to share weights on layer blocks consisting of two layers, or three layers, or all layers ($\pi=2$, or 3 , ...), and use the tuned $\pi$ on test data.

But a uniform sharing strategy might not be optimal because we need to control the degree of sharing in difference levels of the network. For example, for the case in Figure \ref{figure:js}(a), a good choice is to share weights for layers 2-6 and leave layer 1 as it is. Here we present a method that learns the sharing strategy in a dynamic way. To do this, we choose $ln(2) -$ JS divergence as the measure of the similarity between weights of layer $i$ and layer $j$ (denoted as $\mu(i,j)$). Given a layer block ranging from layer $m$ to layer $n$ (denoted as $b(m, n)$), the similarity over the block is defined as

\begin{equation}
sim(m,n) = \frac{\sum_{i=m}^n \sum_{j=m}^n (1-\delta(i,j)) \mu(i,j)}{(n - m + 1) \cdot (n - m)} \label{math-sim-block}
\end{equation}

\noindent where $n - m + 1$ is the size of the block, and $\delta(i,j)$ is the Kronecker delta function. $sim(m,n)$ measures how the weight of a layer is similar to that of another layer in block $b(m,n)$. We can do sharing when $sim(m,n) \ge \theta$ where $\theta$ is the parameter that controls how often a layer is shared.

We begin with layer 1 and search for the biggest block that satisfies the criterion. This process repeats until all the layers are considered, resulting in $N$ layer blocks. For simplicity, we use \{$\pi_1$, ...,$\pi_N$\} (or \{$\pi_i$\}) to represent the blocks in a bottom-up manner (call it sharing policy), where $\pi_i$ is the size of block $i$. Obviously, for an $M$-layer stack we have $\sum_{i=1}^{N}\pi_i = M$.

Once we have a sharing policy, we need to re-train the MT model. It in turn leads to new attention weights and possibly a better policy. A desirable way is to continue learning until the model converges. To this end, we present a joint learning method that trains MT models and sharing policies simultaneously (Figure \ref{fig:joint-learning}). In the method, MT training and policy learning loops for iterations, and the result of the final round is returned when there is no new update of the model.

\begin{figure}
\begin{tabular}{l}
1: \textbf{Function} \textsc{LearnToShare} ($layers$, $model$) \\
2: \hspace{1em} \textbf{while} policy $\{\pi_i\}$ does change \textbf{do} \\
3: \hspace{2em} learn a new $model$ given policy $\{\pi_i\}$ \\
4: \hspace{2em} learn a new policy $\{\pi_i\}$ on $layers$ given $model$ \\
5: \hspace{1em} \textbf{return} $\{\pi_i\}$ \& $model$
\end{tabular}
\caption{\fontsize{9pt}{10.8pt}\selectfont Joint learning of MT models and sharing policies}
\label{fig:joint-learning}
\end{figure}

\section{Experiments}

We experimented with our approach on WMT and NIST translation tasks.

\subsection{Experimental Setup}

The bilingual and evaluation data came from three sources

\begin{itemize}
\item WMT14 (En-De). We used all bilingual data provided within the WMT14 English-German task. We chose newstest 2013 as the tuning data, and newstest 2014 as the test data.
\item WMT17 (En-De, De-En, En-Fi, Fi-En, En-Lv, Lv-En, En-Ru and Ru-En). We followed the standard data setting of the bidirectional translation tasks of German-English, Finnish-English, Latvian-English, and Russian-English. For tuning, we concatenated the data of newstest 2014-2016. For test, we chose newstest 2017.
\item NIST12 (Zh-En). We also used parts of the bitext of NIST OpenMT12 to train a Chinese-English system\footnote{LDC2000T46, LDC2000T47, LDC2000T50, LDC2003E14, LDC2005T10, LDC2002E18, LDC2007T09 and LDC2004T08}. The tuning and test sets were MT06 and MT08.
\end{itemize}

\begin{table}
\setlength{\tabcolsep}{2.8pt}
  \small{
  \centering
  \begin{tabular}{r|c|r|r|r|r|r|r}
  \hline
  \multicolumn{1}{c|}{\multirow{2}{*}{\centering Source}} & \multicolumn{1}{c|}{\multirow{2}{*}{\centering Lang.}} & \multicolumn{2}{c|}{Train} & \multicolumn{2}{c|}{Tune} & \multicolumn{2}{c}{Test} \\ \cline{3-8}
  \multicolumn{1}{c|}{} & \multicolumn{1}{c|}{} & \multicolumn{1}{c|}{sent.} & \multicolumn{1}{c|}{word} & \multicolumn{1}{c|}{sent.}& \multicolumn{1}{c|}{word} & \multicolumn{1}{c|}{sent.} & \multicolumn{1}{c}{word} \\
  \hline
  WMT14 & En-De & 4.5M & 225M & 3000 & 130K & 3003 & 133K \\
  \hline
  \multicolumn{1}{c|}{\multirow{8}{*}{\centering WMT17}} & En-De & \multicolumn{1}{r|}{\multirow{2}{*}{\centering 5.9M}} & \multicolumn{1}{r|}{\multirow{2}{*}{\centering 276M}} & \multicolumn{1}{r|}{\multirow{2}{*}{\centering 8171}} & \multicolumn{1}{r|}{\multirow{2}{*}{\centering 356K}} & 3004 & 128K \\
  & De-En &  &  &  &  & 3004 & 128K \\
  \cline{2-8}
  & En-Fi & \multicolumn{1}{r|}{\multirow{2}{*}{\centering 2.6M}} & \multicolumn{1}{r|}{\multirow{2}{*}{\centering 108M}} & \multicolumn{1}{r|}{\multirow{2}{*}{\centering 8870}} & \multicolumn{1}{r|}{\multirow{2}{*}{\centering 330K}} & 3002 &110K \\
  & Fi-En &  &  &  &  & 3002 & 110K \\
  \cline{2-8}
  & En-Lv & \multicolumn{1}{r|}{\multirow{2}{*}{\centering 4.5M}} & \multicolumn{1}{r|}{\multirow{2}{*}{\centering 115M}} & \multicolumn{1}{r|}{\multirow{2}{*}{\centering 2003}} & \multicolumn{1}{r|}{\multirow{2}{*}{\centering 90K}} & 2001 & 88K \\
  & Lv-En &  &  &  &  & 2001 & 88K \\
  \cline{2-8}
  & En-Ru & \multicolumn{1}{r|}{\multirow{2}{*}{\centering 25M}} & \multicolumn{1}{r|}{\multirow{2}{*}{\centering 1.2B}} & \multicolumn{1}{r|}{\multirow{2}{*}{\centering 8819}} & \multicolumn{1}{r|}{\multirow{2}{*}{\centering 391K}} & 3001 & 132K \\
  & Ru-En &  &  &  &  & 3001 & 132K\\
  \hline
  NIST12 & Zh-En & 1.9M & 85M & 1164 & 227K & 1357 & 198K \\
  \hline
  \end{tabular}
  \caption{\fontsize{9pt}{10.8pt}\selectfont Data statistics (\# of sentences and \# of words)}
  \label{tab:data}
  }
\end{table}

For Chinese, all sentences were word segmented by the segmentation system in the NiuTrans toolkit \cite{xiao-etal-2012-niutrans}. For other languages, we ran the official script of WMT for tokenization.  All sentences of more than 50 words were removed for the NIST Zh-En task, and sentences of more than 80 words were removed for the WMT tasks. For all these tasks, sentences were encoded using byte-pair encoding, where we used a shared source target vocabulary of 32K tokens. See Table \ref{tab:data} for statistics of the data.

We used standard implementation of Transformer. Early versions of its inference system simply compute the attention output for target positions individually. This way is straightforward but with a double counting problem. For a stronger baseline, we chose the system with an attention cache that kept the attention output of previous positions in cache and re-used it in following steps.

The Transformer system used in our experiments consisted of a 6-layer encoder and a 6-layer decoder. By default, we set $d_k = d_v = 512$ and used 2,048 hidden units in the FFN sub-layers. We used multi-head attention (8 heads) because it was shown to be effective for state-of-the-art performance \cite{vaswani2017attention}. Dropout ($\textrm{rate}= 0.1$) and label smoothing ($\epsilon_{ls}=0.1$) methods were adopted for regularization and stabilizing the training \cite{Christian2016Rethinking}. We trained the model using Adam with $\beta_1=0.9$, $\beta_2=0.98$, and $\epsilon=10^{-9}$ \cite{kingma2014adam}. The learning rate was scheduled as described in \cite{vaswani2017attention}: $ lr=d^{-0.5} \cdot min(t^{-0.5}, t \cdot \textrm{4k}^{-1.5})$, where $t$ is the step number. All models were trained for 100k steps with a mini-batch of 4,096 tokens on machines with 8 Nvidia 1080Ti GPUs except En-Fi (60k steps), Fi-En (60k steps) and Zh-En (24k steps). Every model was ensembled from the 5 latest checkpoints in training. For inference, both beam search and batch decoding methods were used (beam size = 4 and batch size = 16).

For our approach, we applied \textsc{San} to self-attention and encoder-decoder sub-layers on the decoder side. We learned sharing policies as in Figure \ref{fig:joint-learning}. $\theta$ was tuned on the tuning data, which resulted in an optimal range of $[0.3,0.4]$ for self-attention and $[0.4,0.5]$ for encoder-decoder attention.

\subsection{Results}

\begin{table}[]
\setlength{\tabcolsep}{2.8pt}
  \small{
  \centering
  \begin{tabular}{l|c|l|r|r|r|r}
    \hline
    \multicolumn{1}{c|}{Source} & \multicolumn{1}{c|}{Language} & \multicolumn{1}{c|}{Model} & \multicolumn{1}{c|}{BLEU} & \multicolumn{1}{c|}{$\Delta_{\textrm{BLEU}}$} & \multicolumn{1}{c|}{Speed}& \multicolumn{1}{c}{$\Delta_{\textrm{Speed}}$} \\
    \hline
    \multicolumn{1}{c|}{\multirow{2}{*}{\centering WMT14}} & \multirow{2}{*}{En-De} & Baseline & 27.52 & 0.00 & 1.03K & 0.00\% \\
    \cline{3-7}
    & & \textsc{San} & 27.69 & +0.17 & 1.44K & +39.81\% \\
    \hline
    \multicolumn{1}{c|}{\multirow{16}{*}{\centering WMT17}} & \multirow{2}{*}{En-De} & Baseline & 28.90 & 0.00 & 1.03K & 0.00\% \\
    \cline{3-7}
    & & \textsc{San} & 28.82 & -0.08 & 1.43K & +38.82\% \\
    \cline{2-7}
    & \multirow{2}{*}{De-En} & Baseline & 34.57 & 0.00 & 0.99K & 0.00\% \\
    \cline{3-7}
    & & \textsc{San} & 34.75 & +0.18 & 1.38K & +39.19\% \\
    \cline{2-7}
    & \multirow{2}{*}{En-Fi} & Baseline & 21.80 & 0.00 & 1.02K & 0.00\% \\
    \cline{3-7}
    & & \textsc{San} & 21.45 & -0.35 & 1.36K & +33.82\% \\
    \cline{2-7}
    & \multirow{2}{*}{Fi-En} & Baseline & 24.94 & 0.00 & 1.02K & 0.00\% \\
    \cline{3-7}
    & & \textsc{San} & 25.25 & +0.31 & 1.25K & +23.28\% \\
    \cline{2-7}
    & \multirow{2}{*}{En-Lv} & Baseline & 15.80 & 0.00 & 0.94K & 0.00\% \\
    \cline{3-7}
    & & \textsc{San} & 16.08 & +0.28 & 1.31K & +39.01\% \\
    \cline{2-7}
    & \multirow{2}{*}{Lv-En} & Baseline & 18.06 & 0.00 & 0.92K & 0.00\% \\
    \cline{3-7}
    & & \textsc{San} & 17.97 & -0.09 & 1.26K & +36.28\% \\
    \cline{2-7}
    & \multirow{2}{*}{En-Ru} & Baseline & 29.93 & 0.00 & 1.02K & 0.00\% \\
    \cline{3-7}
    & & \textsc{San} & 29.51 & -0.42 & 1.29K & +26.17\% \\
    \cline{2-7}
    & \multirow{2}{*}{Ru-En} & Baseline & 33.63 & 0.00 & 1.01K & 0.00\% \\
    \cline{3-7}
    & & \textsc{San} & 33.36 & -0.27 & 1.23K & +21.17\% \\
    \hline
    \multicolumn{1}{c|}{\multirow{2}{*}{\centering NIST12}} & \multirow{2}{*}{Zh-En} & Baseline & 38.59 & 0.00 & 0.84K & 0.00\% \\
    \cline{3-7}
    & & \textsc{San} & 38.19 & -0.40 & 1.02K & +21.34\% \\
    \hline
    \multicolumn{2}{c|}{\multirow{2}{*}{Average}} & Baseline & 27.37 & 0.00 & \multicolumn{1}{c|}{0.98K} & 0.00\% \\
    \cline{3-7}
    \multicolumn{1}{c}{} & \multicolumn{1}{c|}{} & \textsc{San} & 27.31 & -0.07 & \multicolumn{1}{c|}{1.30K} & +31.98\% \\
    \hline
  \end{tabular}
  \caption{\fontsize{9pt}{10.8pt}\selectfont BLEU scores (\%) and translation speeds (token/sec) on the WMT and NIST tasks}
  \label{tab:main-result}
  }
\end{table}

We report the translation quality (in BLEU[\%]) and speed (in token/sec) on all ten of the tasks (Table \ref{tab:main-result}). We see, first of all, that \textsc{San} significantly improves the translation speed for all these languages. The average speed-up is 1.3X. Also, there is a very modest BLEU decrease, but not significant. These results indicate that \textsc{San} is robust and can improve a strong baseline on a wide range of translation tasks. Another interesting finding here is that the speed improvement on En-Ru, Ru-En and Zh-En is not as large as that on other language pairs. This is because we share fewer layers (i.e., larger $\theta$) on these tasks to preserve good BLEU scores. Note that Russian and Chinese are very difficult languages for translation, and we need a complicated network to model the structure divergence. Less sharing is preferred to keep the expressive power for these languages.

\begin{table}[]
\setlength{\tabcolsep}{3pt}
  \small{
  \centering
  \begin{tabular}{l|c|c|r|r|r|r}
    \hline
    \multicolumn{1}{c|}{\multirow{2}{*}{\centering Model}} & \multicolumn{2}{c|}{$\theta$} & \multirow{2}{*}{\centering BLEU} & \multirow{2}{*}{\centering $\Delta_{\textrm{BLEU}}$} & \multirow{2}{*}{\centering Speed}& \multicolumn{1}{c}{\multirow{2}{*}{\centering $\Delta_{\textrm{Speed}}$}} \\
    \cline{2-3}
    \multicolumn{1}{c|}{} & \multicolumn{1}{c|}{\ \ \ \ Self \ \ \ \ } & \multicolumn{1}{c|}{Enc-Dec} & \multicolumn{1}{c|}{} & \multicolumn{1}{c|}{} & \multicolumn{1}{c|}{} & \multicolumn{1}{c}{} \\
    \hline
    Baseline & \multicolumn{2}{c|}{N/A} & 27.52 & 0.00 & 1.03K & 0.00\% \\
    \hline
    \multirow{5}{*}{\centering \textsc{San}} & \multicolumn{2}{c|}{uniform $\{\pi_i\}$} & 27.58 & +0.06 & 1.43K & +38.83\% \\
    \cline{2-7}
    & 0.30 & 0.40 & 26.89 & -0.63 & 1.55K & +50.26\% \\
    \cline{2-7}
    & 0.30 & 0.50 & 27.69 & +0.17 & 1.44K & +39.81\% \\
    \cline{2-7}
    & 0.40 & 0.40 & 26.96 & -0.56 & 1.52K & +47.32\% \\
    \cline{2-7}
    & 0.40 & 0.50 & 27.46 & -0.06 & 1.40K & +36.33\% \\
    \hline
  \end{tabular}
  \caption{\fontsize{9pt}{10.8pt}\selectfont BLEU scores (\%) and translation speeds (token/sec) for different sharing policies. Self = self-attention. Enc-Dec = encoder-decoder attention}
  \label{tab:pi}
  }
\end{table}

\begin{table}[]
\setlength{\tabcolsep}{5pt}
  \normalsize{
  \centering
  \begin{tabular}{l|c|r|r|r|r}
    \hline
    \multicolumn{1}{c|}{Model} & \multicolumn{1}{c|}{Shared $V$} & \multicolumn{1}{c|}{BLEU} & \multicolumn{1}{c|}{$\Delta_{\textrm{BLEU}}$} & \multicolumn{1}{c|}{Speed}& \multicolumn{1}{c}{$\Delta_{\textrm{Speed}}$} \\
    \hline
    Baseline & - & 27.52 & 0.00 & 1.03K & 0.00\% \\
    \hline
    \multicolumn{1}{l|}{\multirow{2}{*}{\centering \textsc{San}}} & no & \multirow{2}{*}{27.49} & \multirow{2}{*}{-0.03} & 1.14K & +10.96\% \\
    \cline{2-2}
    \cline{5-6}
    & yes &  &  & 1.27K & +22.91\% \\
    \hline
  \end{tabular}
  \caption{\fontsize{9pt}{10.8pt}\selectfont BLEU scores (\%) and translation speeds (token/sec) with/without a shared context ($V$) for encoder-decoder layers}
  \label{tab:context}
  }
\end{table}

To modulate the degree of sharing, we study the system behavior under different settings of $\theta$ (Table \ref{tab:pi}) . For comparison, we report the result of uniform $\{\pi_i\}$. Due to the limited space, we present the result on the WMT14 En-De task in the following sensitivity analysis. For uniform sharing, $\{\pi_i\}$ is set to \{6\} for self-attention and \{3,3\} for encoder-decoder attention. This results in a promising speed-up (see entry of uniform $\{\pi_i\}$). When we switch to joint learning of MT models and sharing policies, we obtain further improvements in both BLEU and speed. More interestingly, we see that the system prefers a smaller $\theta$ for self-attention than the encoder-decoder counterpart. This is reasonable because the encoder-decoder attention captures the correspondence of two languages and needs more states in modeling than a single language. On the other hand, the BLEU improvements indicate that the MT system can benefit from simplified models. It gives a direction that we explore simpler models for better training of neural MT systems.

In encoder-decoder attention, we share the context $V$ generated by the encoder for further speed-ups (see Figure \ref{fig:san}(c)). It is therefore worth a study on how much this method can accelerate the system. Table \ref{tab:context} shows that sharing the context contributes half of the speed improvement. This agrees with our design that weight sharing is more beneficial to the decoder because attention is heavier on the decoder side. Another interesting question is whether \textsc{San} can improve the system on the encoder side. To seek an answer, we apply \textsc{San} to the encoder-side self-attention sub-layers and see small speed improvements (Table \ref{tab:encoder-side}). This result confirms the previous report that the decoder occupies the inference time and the encoder is light \cite{Zhang2018AAN}.

\begin{table}[t]
\setlength{\tabcolsep}{8pt}
  \normalsize{
  \centering
  \begin{tabular}{l|r|r|r|r}
    \hline
    \multicolumn{1}{c|}{Model} & \multicolumn{1}{c|}{BLEU} & \multicolumn{1}{c|}{$\Delta_{\textrm{BLEU}}$} & \multicolumn{1}{c|}{Speed}& \multicolumn{1}{c}{$\Delta_{\textrm{Speed}}$} \\
    \hline
    Baseline & 27.52 & 0.00 & 1.03K & 0.00\% \\
    \hline
    \multicolumn{1}{l|}{\centering \textsc{San}} & 27.51 & -0.01 & 1.05K & +1.94\% \\
    \hline
  \end{tabular}
  \caption{\fontsize{9pt}{10.8pt}\selectfont BLEU scores (\%) and translation speeds (token/sec) of the systems that use \textsc{San} on the encoder side}
  \label{tab:encoder-side}
  }
\end{table}

\begin{table}[t]
  \small{
  \centering
  \begin{tabular}{l|r|r|r|r}
    \hline
    \multicolumn{1}{c|}{\multirow{1}{*}{\centering Model}} & \multicolumn{1}{c|}{BLEU} & \multicolumn{1}{c|}{$\Delta_{\textrm{BLEU}}$} & \multicolumn{1}{c|}{Speed}& \multicolumn{1}{c}{$\Delta_{\textrm{Speed}}$} \\
    \cline{2-5}
    \hline
    Baseline & 27.52 & 0.00 & 1.03K & 0.00\% \\
    \hline
    Baseline-Cache & 27.52 & 0.00 & 0.12K & -88.65\% \\
    \hline
    Baseline+\textsc{Aan} & 27.51 & -0.01 & 1.38K & +34.37\% \\
    \hline
    Baseline+\textsc{San} & 27.69 & +0.17 & 1.44K & +39.81\% \\
    \hline
    Baseline+\textsc{Aan}+\textsc{San} & 27.19 & -0.33 & 1.87K & +81.61\% \\
    \hline
  \end{tabular}
  \caption{\fontsize{9pt}{10.8pt}\selectfont Comparison of different attention models}
  \label{tab:comp-model}
  }
\end{table}

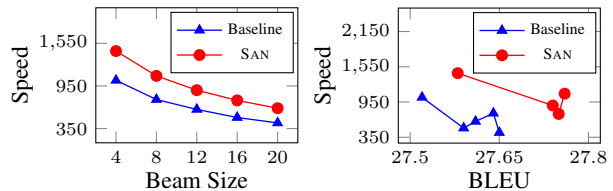
\begin{figure}[t]
\centering
\begin{tabular}{p{3.6cm}p{3.6cm}}
\begin{tikzpicture}{baseline}
\footnotesize{
\begin{axis}[
    width=.24\textwidth,
    height=.19\textwidth,
    legend style={at={(0.98,0.97)},anchor=north east,font=\tiny,legend columns=1,inner sep=2pt},
    xlabel={\footnotesize{Beam Size}},ylabel={\footnotesize{Speed}},
    ylabel style={yshift=-0.8em},xlabel style={yshift=0.8em},
    ymin=150,ymax=2050, ytick={350, 950, 1550},
    xmin=2,xmax=22,xtick={4, 8, 12, 16, 20},xticklabel style={font=\scriptsize},yticklabel style={font=\scriptsize}
]
\addplot[blue,mark=triangle*,line width=0.5pt] coordinates {(4,1028.81) (8,759.07) (12,619.62) (16,505.82) (20,429.90)};
\addlegendentry{Baseline}
\addplot[red,mark=otimes*,line width=0.5pt] coordinates {(4,1440.92) (8,1091.82) (12,890.23) (16,746.66) (20,636.70)};
\addlegendentry{\textsc{San}}	
\end{axis}
}
\end{tikzpicture}
&
\begin{tikzpicture}{baseline}
\footnotesize{
\begin{axis}[
    width=.24\textwidth,
    height=.19\textwidth,
    legend style={at={(0.98,0.97)},anchor=north east,font=\tiny,legend columns=1,inner sep=2pt},
    xlabel={\footnotesize{BLEU}},ylabel={\footnotesize{Speed}},
    ylabel style={yshift=-0.8em},xlabel style={yshift=0.8em},
    ymin=250,ymax=2550, ytick={350, 950, 1550, 2150},
    xmin=27.48,xmax=27.82,xtick={27.50, 27.65, 27.80},xticklabel style={font=\scriptsize},yticklabel style={font=\scriptsize}
]
\addplot[blue,mark=triangle*,line width=0.5pt] coordinates {(27.52,1028.8) (27.59,505.8) (27.61,619.6) (27.64,759.1) (27.65,429.9)};
\addlegendentry{Baseline}
\addplot[red,mark=otimes*,line width=0.5pt] coordinates {(27.58,1440.9) (27.74,890.2) (27.75,746.7) (27.76,1091.8)};
\addlegendentry{\textsc{San}}	
\end{axis}
}
\end{tikzpicture}
\\
\end{tabular}
\caption{\fontsize{9pt}{10.8pt}\selectfont Translation speed (token/sec) vs beam size and BLEU (\%)}
\label{fig:beam-size}
\end{figure}

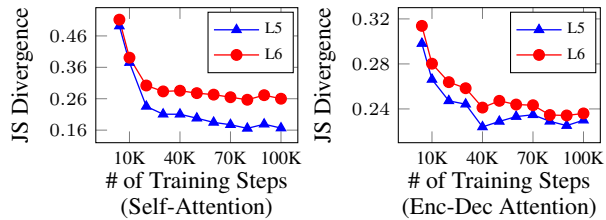
\begin{figure}[t]
\centering
\begin{tabular}{p{3.6cm}p{3.6cm}}
\begin{tikzpicture}{baseline}
\footnotesize{
\begin{axis}[
    width=.24\textwidth,
    height=.19\textwidth,
    legend style={at={(0.98,0.97)},anchor=north east,font=\tiny,legend columns=1,inner sep=2pt},
    ylabel style={yshift=-0.8em},xlabel style={yshift=0.8em,align=center},
    xlabel={\footnotesize{\# of Training Steps}\\\footnotesize{(Self-Attention)}},ylabel={\footnotesize{JS Divergence}},
    ymin=0.12,ymax=0.55, ytick={0.16, 0.26, 0.36, 0.46},
    xmin=-10,xmax=110,xtick={10, 40, 70, 100}, xticklabels={10K, 40K, 70K, 100K},xticklabel style={font=\scriptsize},yticklabel style={font=\scriptsize}
]
\addplot[blue,mark=triangle*,line width=0.5pt] coordinates {(4, 0.4915) (10,0.3752) (20,0.2356) (30,0.2107) (40,0.2106) (50,0.1980) (60,0.1846) (70,0.1773) (80,0.1656) (90,0.1789) (100,0.1666)};
\addlegendentry{L5}
\addplot[red,mark=otimes*,line width=0.5pt] coordinates {(4, 0.5118) (10,0.3907) (20,0.3021) (30,0.2833) (40,0.2855) (50,0.2778) (60,0.2730) (70,0.2648) (80,0.2567) (90,0.2713) (100,0.2600)};
\addlegendentry{L6}
\end{axis}
}
\end{tikzpicture}
&
\begin{tikzpicture}{baseline}
\footnotesize{
\begin{axis}[
    width=.24\textwidth,
    height=.19\textwidth,
    legend style={at={(0.98,0.97)},anchor=north east,font=\tiny,legend columns=1,inner sep=2pt},
    ylabel style={yshift=-0.8em},xlabel style={yshift=0.8em, align=center},
    xlabel={\footnotesize{\# of Training Steps}\\\footnotesize{(Enc-Dec Attention)}},ylabel={\footnotesize{JS Divergence}},
    ymin=0.21,ymax=0.33, ytick={0.24, 0.28, 0.32},
    xmin=-10,xmax=110,xtick={10, 40, 70, 100},, xticklabels={10K, 40K, 70K, 100K}, xticklabel style={font=\scriptsize},yticklabel style={font=\scriptsize}
]
\addplot[blue,mark=triangle*,line width=0.5pt] coordinates {(4, 0.2981) (10,0.2662) (20,0.2472) (30,0.2442) (40,0.2240) (50,0.2289) (60,0.2332) (70,0.2348) (80,0.2290) (90,0.2252) (100,0.2300)};
\addlegendentry{L5}
\addplot[red,mark=otimes*,line width=0.5pt] coordinates {(4, 0.3138) (10,0.2802) (20,0.2638) (30,0.2584) (40,0.2412) (50,0.2472) (60,0.2439) (70,0.2433) (80,0.2346) (90,0.2342) (100,0.2360)};
\addlegendentry{L6}
\end{axis}
}
\end{tikzpicture}
\\
\end{tabular}
\caption{\fontsize{9pt}{10.8pt}\selectfont JS divergence vs number of training steps}
\label{fig:JS-training-steps}
\end{figure}

Also, we plot the translation speed as functions of beam size and BLEU score. Figure \ref{fig:beam-size} shows a consistent improvement under different beam settings. Moreover, \textsc{San} benefits more from larger beam sizes. For example, the speed-up of beam = 20 is larger than that of beam = 4 (1.48x vs. 1.40x). The Speed-vs-BLEU curves indicate a good ability of \textsc{San} in trading off between translation quality and speed.

In addition, we empirically compare \textsc{San} with other variants of the attention model, including \textsc{Aan} \cite{Zhang2018AAN} and the model with no cache. Table \ref{tab:comp-model} shows the attention cache plays an important role in fast inference. It leads to an 8-fold speed-up on top of the implementation where no cache is used. Also, \textsc{San} obtains a bigger speed improvement than \textsc{Aan}. This might be because \textsc{Aan} is used for self-attention only, while \textsc{San} is applicable to both self-attention and encoder-decoder attention. Finally, we combine \textsc{Aan} and \textsc{San} in a new system where \textsc{Aan} is applied to self-attention and \textsc{San} is applied to encoder-decoder attention. It yields the best result which is 1.8 times faster than the baseline, and almost 16 times faster than the system without cache.

In training, we observe that systems tend to learn similar attention weights. Figure \ref{fig:JS-training-steps} plots the JS divergence between layer 4 and layers 5-6 at different training steps. The JS divergence curves go down significantly as the training proceeds. Adjacent layers show more similar weight distributions than distant layers. The fast convergence in JS divergence can speed up the iterative training. For example, for each training epoch (Figure \ref{fig:joint-learning}), one can train the model for a shorter time, as the JS divergence among layers converges quickly. Thus, the system can find the optimal sharing policy more efficiently. In addition, we find that the training likelihood of \textsc{San} is higher than that of the baseline, but not significant.

\section{Related Work}
It has been observed that attention models are critical for state-of-the-art results on many MT tasks \cite{bahdanau2014neural,wu2016google,vaswani2017attention}. Several research groups have investigated attentive methods for different architectures of neural MT. The earliest is \cite{luong2015effective}. They introduced an additive attention model into MT systems based on recurrent neural networks (RNNs). More recently, multi-layer attention was successfully applied to convolutional neural MT systems \cite{gehring2017convs2s} and Transformer systems \cite{vaswani2017attention}. In particular, Transformer is popular due to its scalability on large-scale training and the good design of the architecture for implementation.

It is well-known that Transformer suffers from a high inference cost which makes it slower than the RNN-based counterpart. This partially due to the auto-regressive property of decoding, and partially due to the heavy use of dot-product attention where the expensive matrix multiplication is frequently used. Researchers have begun to explore solutions. For example, Gu et al. \shortcite{Gu2017NonAutoregressive} designed a non-autoregressive inference method for a Transformer-like system, which generated the entire target sequence at one time. This model is fast but is not easy to train.

In another line of research, Zhang et al. \shortcite{Zhang2018AAN} proposed the average attention network (\textsc{Aan}) and applied it to self-attention sub-layers on the decoder side with no cache. In this work, we study the issue on a strong baseline that has already used an attention cache for a reasonable inference speed. Also, our approach is straightforwardly applicable to systems with multi-layer attention. We improve both the self-attention and encoder-decoder attention components, which has not been studied in previous work.

In neural MT, fast inference methods have been investigated for years. These include vocabulary selection \cite{Gurvan2016Selection,Baskaran2017Selection}, knowledge distillation \cite{Hinton2015Distilling,Yoon2016Distillation}, low-precision computation \cite{Paulius2017Precision,Jerry2018EightBit}, recurrent stacked networks \cite{Dabre2019RecurrentSO} and etc. Our method is orthogonal to them. Previous studies focus more on the reduction of model size and robust training, rather than fast inference. Here we study the issue in the context of speeding up attentive MT and confirm the effectiveness of this kind of models.

\section{Conclusions}

We have presented a shared attention network (\textsc{San}) for fast inference of Transformer. It shares attention weights among layers for both self-attention and encoder-decoder attention in a vertical manner. The policy of sharing can be jointly learned with the MT model, rather than being determined heuristically. Moreover, \textsc{San} reduces the memory footprint. Experiments on ten MT tasks show that \textsc{San} yields a speed-up of 1.3X over a strong baseline that has already used an attention cache. More interestingly, it is observed that the combination of \textsc{San} and \textsc{Aan} obtains a larger speed improvement. The system is 16X faster than the baseline with no cache.

\section*{Acknowledgments}

This work was supported in part by the National Science Foundation of China (Nos. 61732005, 61876035 and 61432013) and the Fundamental Research Funds for the Central Universities (No. N181602013).


\bibliographystyle{ijcai19}
\bibliography{ijcai19}

\end{document}